\documentclass[letterpaper, 10 pt, journal]{IEEEtran}  

\IEEEoverridecommandlockouts                              

\usepackage{graphics} 
\usepackage{amsmath} 
\usepackage{amssymb}  
\usepackage[makeroom]{cancel}
\usepackage{colortbl}
\usepackage{pifont}
\usepackage{mdframed}
\usepackage{balance}
\usepackage{multirow}
\usepackage{algorithm}
\usepackage{algpseudocode}

\newcommand{\cmark}{\ding{51}}%
\newcommand{\xmark}{\ding{55}}%

\newcommand{\hide}[1]{}

\newcommand{\bdmath}{\begin{dmath}}
\newcommand{\edmath}{\end{dmath}}
\newcommand{\beq}{\begin{equation}}
\newcommand{\eeq}{\end{equation}}
\newcommand{\bdm}{\begin{displaymath}}
\newcommand{\edm}{\end{displaymath}}
\newcommand{\bea}{\begin{eqnarray}}
\newcommand{\eea}{\end{eqnarray}}
\newcommand{\beal}{\beq \begin{array}{ll}}
\newcommand{\eeal}{\end{array} \eeq}
\newcommand{\beas}{\begin{eqnarray*}}
\newcommand{\eeas}{\end{eqnarray*}}
\newcommand{\ba}{\begin{array}}
\newcommand{\ea}{\end{array}}
\newcommand{\bit}{\begin{itemize}}
\newcommand{\eit}{\end{itemize}}
\newcommand{\ben}{\begin{enumerate}}
\newcommand{\een}{\end{enumerate}}

\newcommand{\SO}{\mathrm{SO}}
\newcommand{\so}{\mathfrak{so}}

\newcommand{\SOthree}{\ensuremath{\SO(3)}\xspace}
\newcommand{\sothree}{\ensuremath{\so(3)}\xspace}

\newcommand{\R}{\mathtt{R}}

\newcommand{\rotvel}{\boldsymbol\omega}
\newcommand{\acc}{\mathbf{a}}

\newcommand{\tran}{\mathbf{t}}

\newcommand{\vel}{\mathbf{v}}

\newcommand{\bias}{\mathbf{b}}

\newcommand{\gravity}{\mathbf{g}}
\newcommand{\noise}{\boldsymbol\eta}

\newcommand{\expmap}{\mathrm{Exp}}
\newcommand{\logmap}{\mathrm{Log}}

\usepackage[dvipsnames]{xcolor}

\usepackage{amsmath}
\usepackage{amssymb}
\usepackage{amsthm}
\usepackage{mathtools}
\usepackage{xspace}

\usepackage{graphicx}
\usepackage{subcaption}
\usepackage[percent]{overpic}
\usepackage{booktabs}
\usepackage{tabularray}
\UseTblrLibrary{booktabs}
\usepackage{rotating}
\usepackage{tikz}
\usetikzlibrary{arrows.meta}
\usepackage{mathabx}

\usepackage{ragged2e} 
\usepackage{pifont}
\usepackage{lipsum}

\usepackage{gensymb}
\usepackage{comment}
\usepackage{multirow}
\usepackage{tabularray}
\usepackage[export]{adjustbox}
\usepackage{rotating}
\usepackage{makecell}

\usepackage{algorithm}
\usepackage{algpseudocode}

\AtEndPreamble{
    \usepackage[capitalize]{cleveref}
    \crefname{section}{Sec.}{Secs.}
    \Crefname{section}{Section}{Sections}
    \Crefname{table}{Table}{Tables}
    \crefname{table}{Tab.}{Tabs.}
}

\definecolor{darkgreen}{rgb}{0.0,0.6,0.0}

\usepackage{amssymb} 
\usepackage{breqn}
\usepackage{multirow}
\usepackage{cancel} 
\usepackage{xfrac}
\usepackage{xspace}
\usepackage{booktabs}
\usepackage{pifont}
\usepackage{arydshln}
\usepackage{makecell}  
\usepackage{balance}
\usepackage{verbatim}
\usepackage{comment}
\makeatletter
\let\NAT@parse\undefined
\makeatother
\usepackage[colorlinks=true,
			linkcolor=blue,
			urlcolor=blue,
			citecolor=blue]{hyperref}

\makeatletter
\DeclareRobustCommand\onedot{\futurelet\@let@token\@onedot}
\def\@onedot{\ifx\@let@token.\else.\null\fi\xspace}

\makeatother

\definecolor{pastelred}{RGB}{235,119,119}
\definecolor{pastelorange}{RGB}{255,202,126}

\makeatletter
\newcommand{\setword}[2]{%
  \phantomsection
  #1\def\@currentlabel{\unexpanded{#1}}\label{#2}%
}
\makeatother

\begin{document}

\title{An Efficient Closed-Form Solution to Full Visual-Inertial State Initialization}

\author{Samuel Cerezo$^1$, Seong Hun Lee$^1$, Javier Civera$^1$
\thanks{Manuscript received: November, 24, 2025; Revised February, 2, 2026; Accepted March, 25, 2026.}
\thanks{This paper was recommended for publication by Editor Pascal Vasseur upon evaluation of the Associate Editor and Reviewers' comments.
This work was supported by the Spanish Government (PID2021-127685NB-I00, TED2021-131150B-I00), the Agencia Estatal de Investigación / Ministerio de Ciencia e Innovación (AEI/MCIN) under grant PRE2022-103765, and the Aragón Government
(project T45\_23R).} 
\thanks{$^1$The authors are with the I3A,
        Universidad de Zaragoza, C/María de Luna, 1. 50018, Zaragoza, Spain.
        {\tt\small \{sacerezo,seonghunlee,jcivera\}@unizar.es}}
\thanks{Digital Object Identifier (DOI): see top of this page.}
}

\markboth{IEEE Robotics and Automation Letters. Preprint Version. Accepted March, 2026}
{Cerezo \MakeLowercase{\textit{et al.}}: An Efficient Closed-Form Solution to Full Visual-Inertial State Initialization} 



\maketitle

\begin{abstract}
In this letter, we present a closed-form initialization method that recovers the full visual–inertial state without nonlinear optimization. Unlike previous approaches that rely on iterative solvers, our formulation yields analytical, easy-to-implement, and numerically stable solutions for reliable start-up. Our method builds on small-rotation and constant-velocity approximations, which keep the formulation compact while preserving the essential coupling between motion and inertial measurements. We further propose an observability-driven, two-stage initialization scheme that balances accuracy with initialization latency. Extensive experiments on the EuRoC dataset validate our assumptions: our method achieves $10-20\%$ lower initialization error than optimization-based approaches, while using $4\times$ shorter initialization windows and reducing computational cost by $5\times$.
\end{abstract}

\section{Introduction}

\IEEEPARstart{V}{isual–inertial} SLAM has become a core technology in robotics and mobile devices, enabling real-time localization and mapping by fusing camera and inertial measurements~\cite{qin2018vins,campos2021orb,boche2025okvis2}. 
While camera intrinsics can be calibrated once and remain relatively stable over time, inertial sensors must be re-initialized at each run. 
This initialization step estimates critical quantities such as the initial velocity, gravity direction, and gyroscope and accelerometer biases, which serve as the seed for subsequent state estimation. 
Accurate initialization is therefore essential: poor initial estimates may cause optimization or filtering to converge to local minima, leading to scale drift, degraded tracking, or even complete failure of the system. 
Conversely, proper initialization enhances robustness under challenging conditions and recovers the metric scale that monocular vision alone cannot provide.

A central insight from the SLAM literature is that orientation plays a pivotal role: once rotations are reliably estimated, the recovery of the remaining states becomes more robust. 
Carlone and Censi~\cite{carlone2014angular} showed that accurate orientation estimates prevent spurious minima in pose-graph optimization, and Carlone et al.~\cite{carlone2015initialization} showed that good rotation initialization significantly improves convergence and accuracy in large-scale SLAM. 
These results motivate our approach, which relies on simple but accurate decoupled orientation approximations to estimate the rest of the states.

Existing visual–inertial initialization strategies fall into two main categories. 
Closed-form solutions have been proposed by Martinelli~\cite{martinelli2011vision,martinelli2014closed}, but they assume partial knowledge of the states, specifically of the sensor biases. 
Later works addressed this limitation using iterative nonlinear optimization~\cite{kaiser2016simultaneous,dominguez2018visual,campos2019fast,campos2020inertial}, which achieves accurate full state initialization but requires careful seeding and incurs significant computational cost. 
More recent formulations attempt to reduce complexity by decoupling rotation and translation~\cite{lee2021rotation,concha2021instant,he2023rotation}, or by combining closed-form solutions with filtering schemes~\cite{wang2023edi}. 
However, none of these approaches provides a fully analytical solution for the complete visual–inertial state.

In this letter, we present the first closed-form initialization method for all states of a visual–inertial system, including gyroscope and accelerometer biases. 
Our formulation builds on small-rotation and constant-velocity approximations, which hold for common sensor hardware and motion patterns. We further contribute a two-stage criterion that assesses the observability of the visual–inertial states and enables reliable initialization under diverse conditions.
Extensive experiments show that our method achieves higher accuracy than relevant baselines, while significantly reducing the computational cost and initialization delay.

\section{Related Work}

Early works on closed-form initialization were introduced by Martinelli~\cite{martinelli2011vision,martinelli2014closed}, who derived minimal unique solutions under the assumption of known gyroscope and accelerometer biases. Subsequent approaches shifted towards nonlinear optimization to jointly estimate these biases together with the remaining states~\cite{kaiser2016simultaneous,dominguez2018visual}, and later refinements such as those by Campos et al.~\cite{campos2019fast,campos2020inertial} achieved robust results but still relied on iterative procedures requiring careful seeding.  

More recently, alternative formulations have been proposed to improve efficiency and robustness. Evangelidis and Micušík~\cite{evangelidis2020revisiting} revisited the problem with a new closed-form solver for the vi-SfM problem, elegant but better suited for offline reconstruction. Decoupled approaches to relative~\cite{kneip2013direct} and multi-view~\cite{lee2021rotation} camera motion have also been proposed~\cite{concha2021instant,he2023rotation}, but they still require nonlinear optimization and in some cases cannot recover accelerometer bias. Hybrid strategies such as the framework of Wang et al.~\cite{wang2023edi} can estimate both biases through an error-state Kalman filter, at the expense of additional algorithmic complexity.  

In parallel, approximations have played an important role in related problems: the small-angle approximation in triangulation~\cite{oliensis2002exact,lee2019closed}, and the small-rotation approximation in vision-only reconstruction from accidental motion~\cite{yu20143d} and efficient relative pose solvers~\cite{VenturaAprox2015}. However, to the best of our knowledge, the small-rotation approximation has never been exploited for visual-inertial state initialization.  

In summary, existing methods either assume known biases, rely on iterative optimization, or adopt hybrid formulations that increase complexity (see Table~\ref{tab:relwork}). In contrast, we propose a full-state closed-form solution by only assuming small rotations—a reasonable condition for typical hardware and most use cases. Our approach therefore combines the accuracy of optimization-based methods with the simplicity and efficiency of analytical solvers.

\begin{table}[t]
\centering
\setlength{\tabcolsep}{5pt}
\begin{tabular}{l | c c c}
\toprule
Method & Closed-form & Gyro-bias & Accel-bias  \\
\midrule
Martinelli~\cite{martinelli2014closed} & \color{darkgreen}{\cmark} & \color{red}{\xmark} & \color{red}{\xmark}\\
Kaiser et al.~\cite{kaiser2016simultaneous} & \color{red}{\xmark} & \color{darkgreen}{\cmark} & \color{darkgreen}{\cmark} \\
Domínguez et al.~\cite{dominguez2018visual} & \color{red}{\xmark} & \color{darkgreen}{\cmark} & \color{darkgreen}{\cmark} \\
Campos et al.~\cite{campos2019fast,campos2020inertial} & \color{red}{\xmark} & \color{darkgreen}{\cmark} & \color{darkgreen}{\cmark} \\
Concha et al.~\cite{concha2021instant} & \color{red}{\xmark} & \color{darkgreen}{\cmark} & \color{darkgreen}{\cmark} \\
{Evangelidis and Micušík~\cite{evangelidis2020revisiting}} & {\color{darkgreen}\cmark} & {\color{red}\xmark} & {\color{red}\xmark} \\
{Wang et al.~\cite{wang2023edi}} & {\color{red}\xmark} & {\color{darkgreen}\cmark} & {\color{darkgreen}\cmark} \\
\midrule
\textbf{Ours} & \color{darkgreen}{\cmark} & \color{darkgreen}{\cmark} & \color{darkgreen}{\cmark}\\
\bottomrule
\end{tabular}
\caption{\textbf{Comparison of existing initialization methods.} Summary of whether each method provides a closed-form solution and estimates gyroscope and accelerometer biases.}
\label{tab:relwork}
\end{table}
\section{Notation and Preliminaries}

Vectors are denoted by bold lowercase letters ($\mathbf{x}$), with unit vectors indicated by a hat accent ($\widehat{\mathbf{x}}$). Bold upper-case letters ($\mathbf{A}$) represent matrices, and light lower-case letters ($a$) denote scalars. Rotation matrices $\R_{ab} \in \SOthree$ refer to the transformation from frame $b$ to frame $a$, omitting the first subindex when referring to the world frame $w$, \emph{i.e.}, $\R_{a} \doteq \R_{wa}$. In axis–angle form, $\mathbf{r} = \theta \widehat{\mathbf{r}} \in \mathbb{R}^3$, with the unit rotation axis $\widehat{\mathbf{r}} = \frac{\mathbf{r}}{||\mathbf{r}||} \in \mathbb{S}^2$ and the rotation angle $\theta = ||\mathbf{r}|| \in  \mathbb{R}$. The skew-symmetric matrix of $\mathbf{r}$ is denoted as ${\mathbf{r}}^\wedge$.
The IMU integration between time instants $t_i$ and $t_j$ follows the standard formulation~\cite{On-manifold2017}
\begin{align}
	\R_j
    =&\ \R_i
    \prod\nolimits_{k=i}^{j-1}\expmap\left(\left(\tilde\rotvel_k - \bias^g_k-\noise^{gd}_k \right) \Delta t\right),  \label{eq:preintegrationRotation}\\
  \vel_j
    =&\ \!  \vel_i \!  + \gravity \Delta t_{ij}
    + \sum\nolimits_{k=i}^{j-1} \R_k \Big( \tilde\acc_k-\bias^a_k-\noise^{ad}_k \Big)\Delta t \label{eq:preintegrationVelocity}\\ 
  \tran_j
    =&\ \! \tran_i \!
    + \sum_{k=i}^{j-1}
      \Big[ \vel_k \Delta t
    + \frac{1}{2}\gravity \Delta t^2
    + \frac{1}{2}\R_k \Big( \tilde\acc_k \!-\! \bias^a_k \!-\! \noise^{ad}_k \Big)\Delta t^2 \Big] \label{eq:preintegrationTranslation}
\end{align}
\noindent
where $\vel_i$ and $\vel_j$ denote the linear velocity at times $t_i$ and $t_j$, $\gravity$ the gravity vector, and $\Delta t$ the sampling interval between consecutive measurements. 
$\tilde\rotvel_k$ and $\tilde\acc_k$ are the gyroscope and accelerometer readings at time $k$, with associated biases $\bias^g_k$ and $\bias^a_k$, respectively. 
We assume that gyroscope and accelerometer biases remain approximately constant during initialization, \emph{i.e.}, $\bias^g_k \approx \bias^g$ and $\bias^a_k \approx \bias^a$. We denote the number of preintegrated IMU measurements by $L = j-i-1$, and hence $t_j-t_i = L \Delta t$. 




\section{Rotation and Gyro Bias Initialization}
\label{sec:rot-and-bias-estimation}


For a small ($\theta \rightarrow 0$) rotation vector $\mathbf{r}$, the Rodrigues’ formula relates it to a rotation matrix representation as
\begin{equation}
    \R = \mathbf{I}_3 + \frac{\sin{\theta}}{\theta} {\mathbf{r}}^\wedge + \frac{1 - \cos{\theta}}{\theta^2}   {\mathbf{r}}^{\wedge 2}
    \underset{\theta \rightarrow 0}{\approx} \mathbf{I}_3 + {\mathbf{r}}^\wedge \,,
\end{equation}
\noindent where $\mathbf{I}_3$ is the $3\times 3$ identity matrix.
Notably, under this approximation, rotation matrices commute only up to second-order terms. Approximating $\R_{ab} \approx \mathbf{I} + {\mathbf{r}_{ab}}^\wedge$ and $\R_{cd} \approx \mathbf{I} + {\mathbf{r}_{cd}}^\wedge$, the products $\R_{ab}\R_{cd}$ and $\R_{cd}\R_{ab}$ then become
\begin{eqnarray}
    \R_{ab}\R_{cd} \approx \mathbf{I} +{\mathbf{r}_{ab}}^\wedge + \ \! {\mathbf{r}_{cd}}^\wedge + \cancel{{\mathbf{r}_{ab}}^\wedge \ \! {\mathbf{r}_{cd}}^\wedge} \label{eq:smallrotcomm1} \\
    \R_{cd}\R_{ab} \approx \mathbf{I} +  {\mathbf{r}_{cd}}^\wedge +  {\mathbf{r}_{ab}}^\wedge + \cancel{{\mathbf{r}_{cd}}^\wedge  \ \! {\mathbf{r}_{ab}}^\wedge} \label{eq:smallrotcomm2}
\end{eqnarray}
differing only in the neglected second-order cross products of the skew-symmetric matrices 
${\mathbf{r}_{ab}}^\wedge {\mathbf{r}_{cd}}^\wedge$ and ${\mathbf{r}_{cd}}^\wedge {\mathbf{r}_{ab}}^\wedge$,which generally do not commute\footnote{Skew-symmetric matrices actually commute with their adjoints.}. 
However, under the small-rotation assumption, these second-order contributions are negligible relative to the first-order terms, making the two matrix products approximately equal.

Rodrigues’ formula also provides the exponential map 
$\exp: \sothree \rightarrow  \SOthree$. 
For small rotation vectors $\mathbf{r}_{ab}$ and $\mathbf{r}_{cd}$,  
the exponential of their sum approximates the product of exponentials:
\begin{equation}
\small
\begin{gathered}
\expmap\!\left({\mathbf{r}_{ab}} + {\mathbf{r}_{cd}}\right)
\underset{\theta \rightarrow 0}{\approx}
\mathbf{I}_3 + \left({\mathbf{r}_{ab}} + {\mathbf{r}_{cd}}\right)^\wedge
= \mathbf{I}_3 + {\mathbf{r}_{ab}}^\wedge + {\mathbf{r}_{cd}}^\wedge \\
=
\left(\mathbf{I}_3 + {\mathbf{r}_{ab}}^\wedge\right)
\left(\mathbf{I}_3 + {\mathbf{r}_{cd}}^\wedge\right)
= \expmap\!\left({\mathbf{r}_{ab}}\right)
\expmap\!\left({\mathbf{r}_{cd}}\right)
\end{gathered}
\label{eq:expsumsumexp}
\end{equation}
With this, the exponential maps in Eq.~(\ref{eq:preintegrationRotation}) 
can be decomposed into products as shown below
\begin{equation}
\small
\prod_{k=i}^{j-1}\expmap\!\left((\tilde\rotvel_k - \bias^g) \Delta t\right) 
\underset{ \theta \rightarrow 0}{\approx}\prod_{k=i}^{j-1}\expmap(\tilde\rotvel_k\Delta t)\expmap(- \bias^g \Delta t)
\label{eq:smallanglepreint}
\end{equation}
By first commuting the products of the right part, and then applying Eq.~\eqref{eq:expsumsumexp}, we obtain the following expression
\begin{eqnarray}
    \R_{ij} = \R_i^\top \R_j \underset{ \theta \rightarrow 0}{\approx} \prod\nolimits_{k=i}^{j-1}\left( \expmap\left(\tilde\rotvel_k\Delta t\right)\expmap\left( - \bias^g \Delta t\right) \right) \label{eq:nocommuting} \\ \underset{ \theta \rightarrow 0}{\approx} \prod\nolimits_{k=i}^{j-1}\left( \expmap\left(\tilde\rotvel_k\Delta t\right) \right) \prod\nolimits_{k=i}^{j-1} \left( \expmap\left( - \bias^g \Delta t\right) \right) \\
    = \prod\nolimits_{k=i}^{j-1}\left( \expmap\left(\tilde\rotvel_k\Delta t\right) \right) \ \expmap\left( -L \bias^g \Delta t\right)
\end{eqnarray}
from which $\bias^g$ can be solved analytically. This yields our first closed-form solution, referred to as \textbf{commutative approximation}
\begin{mdframed}[linewidth=0.5pt, roundcorner=1pt,frametitleaboveskip=2pt,frametitlebelowskip=2pt, innertopmargin=0pt, innerbottommargin=3pt, frametitle={\small$\bias^g$ using the commutative approximation},frametitlebackgroundcolor=gray!15]
\begin{equation}
\bias^g \approx -\frac{1}{L \Delta t}\logmap \left( \left(\prod_{k=i}^{j-1}\expmap\left(\tilde\rotvel_k\Delta t\right)\right)^\top \R_{ij} \right)
    \label{eq:bg-commutative-model}
\end{equation}
\end{mdframed}
We can derive more efficient closed-form solutions for $\bias^g$ by assuming constant rotation angles within each IMU integration. 
First, we express the left-hand side of Eq.~(\ref{eq:nocommuting}) as the composition of $L$ identical rotations:
\begin{equation}
\small
\R_{ij} = \prod_{k=i}^{j-1}\expmap\!\left( \tfrac{\logmap(\R_{ij})}{L} \right) 
\underset{ \theta \rightarrow 0}{\approx} 
\prod_{k=i}^{j-1}\left( \expmap(\tilde\rotvel_k\Delta t)\expmap(- \bias^g \Delta t) \right) 
\label{eq:dividerotation}
\end{equation}
Now, suppose that $\tilde\rotvel_k$ remains approximately constant within the time interval between two images. Approximating $\tilde\rotvel_k$ in Eq.~(\ref{eq:dividerotation}) by their average $\overline\rotvel_k$ and dividing both sides of the equation by $L$ identical rotations, we get
\begin{equation}
\expmap\!\left( \frac{\logmap(\R_{ij})}{L} \right) 
\approx 
\expmap(\overline\rotvel_k\Delta t)\expmap(- \bias^g \Delta t) ,
\end{equation}
from which we can also solve for $\bias^g$ analytically. We can compute $\overline\rotvel_k$ by integrating all IMU measurements or by just averaging them arithmetically, which will give us two closed-form solutions that we will denote as \textbf{average approximation} and \textbf{arithmetic average approximation} respectively.


\begin{mdframed}[linewidth=0.5pt, roundcorner=1pt,frametitleaboveskip=2pt,frametitlebelowskip=2pt, innertopmargin=0pt, innerbottommargin=3pt, frametitle={\small$\bias^g$ using the average approximation},frametitlebackgroundcolor=gray!15]
\begin{align}
& \bias^g \approx -\frac{1}{\Delta t} 
\logmap \!\left( \expmap(\overline\rotvel_k\Delta t)^\top \expmap\!\left( \tfrac{\logmap(\R_{ij})}{L} \right)\right) \notag \\
& \text{with} \ \ \  \overline\rotvel_k \approx \frac{1}{L\Delta t}\logmap\!\left(\prod\nolimits_{k=i}^{j-1} \expmap(\tilde\rotvel_k\Delta t)\right) 
\label{eq:const-vel-model1}
\end{align}
\end{mdframed}

\begin{mdframed}[linewidth=0.5pt, roundcorner=1pt,frametitleaboveskip=2pt,frametitlebelowskip=2pt, innertopmargin=0pt, innerbottommargin=3pt, frametitle={\small$\bias^g$ using the arithmetic average approximation},frametitlebackgroundcolor=gray!15]
\begin{align}
& \bias^g \approx -\frac{1}{\Delta t} 
\logmap \!\left( \expmap(\overline\rotvel_k\Delta t)^\top \expmap\!\left( \tfrac{\logmap(\R_{ij})}{L} \right)\right) \notag\\
& \text{with} \ \ \   \overline\rotvel_k = \frac{1}{L} \sum\nolimits_{k=i}^{j-1}\tilde\rotvel_k
\label{eq:const-vel-model2}
\end{align}
\end{mdframed}

$\R_{ij}$ can be estimated from feature correspondences between any pair of frames. However, in practice, it is both feasible and convenient to initialize $\R_{ij}$ and $\bias^g$ using only the first two frames. First, estimation errors tend to grow over time. Inertial drift accumulates, and our approximations become less accurate as more IMU readings are involved.
Second, unlike translation estimates, relative rotation estimates do not necessarily benefit from larger inter-frame motions. In fact, large motions often introduce perspective distortions in feature appearance, reducing the number of reliably tracked points and increasing matching noise~\cite{fontan2022model}.
{Although $\R_{ij}$ is obtained from visual correspondences, our approach does not rely on a full SFM-based pipeline. Rotation is estimated independently of the structure using a certifiably optimal two-view solver, thereby avoiding scale ambiguity and depth-related instabilities. This rotation-only formulation leads to a more robust gyroscope bias estimation than multi-view SFM approaches estimating six-degrees-of-freedom poses and depths. } 

\section{Initial Velocity, Translation, Gravity and Accelerometer Bias}

\subsection{Closed-form initialization from linear point constraints}
Once $\R_{ij}$ and $\bias^g$ are initialized, the relation between inertial measurements and image tracks can be expressed linearly. Let the sets $\mathcal{Z}_i = \{ \mathbf{z}_i^1, \hdots, \mathbf{z}_i^n, \hdots, \mathbf{z}_i^N \}$ and $\mathcal{Z}_j = \{ \mathbf{z}_j^1, \hdots, \mathbf{z}_j^n, \hdots, \mathbf{z}_j^N \}$ denote the image coordinates of $N$ visual features extracted in an image at $t_i$ and tracked up to a later image, not necessarily consecutive, taken at $t_j$. Assuming a pinhole camera model, every feature $\mathbf{z}_c^n \in \Omega \subset \mathbb{R}^2$ in both images $c \in \{ i, j\}$ can be backprojected to a 3D point $\mathbf{p}_c^n \in \mathbb{R}^3$ in the local camera frame as follows
\begin{equation}
\mathbf{p}_c^n = \lambda_c^n \mathtt{K}^{-1} 
\begin{bmatrix} {\mathbf{z}_c^n}^\top & 1 \end{bmatrix}^\top = \lambda_c^n \boldsymbol{\mu}_c^n,
\label{eq:pi-pj-back-prop}
\end{equation}
where $\mathtt{K} \in \mathbb{UT}^{+}(3)$ with $\mathtt{K}[3,3]=1$ is the camera calibration matrix, $\Omega$ is the image domain, $\boldsymbol{\mu}_j^n \in \mathcal{S}^2$ is the normalized feature bearing and $\lambda_j^n \in \mathbb{R}$ its distance along such bearing.
Following~\cite{martinelli2014closed}, a linear constraint can be derived for every feature $\mathbf{z}_c^n$ in every pair of frames taken at $t_i$ and $t_j$
\vspace{-3pt}
\begin{equation}
\small
\begin{split}
-\frac{(L\Delta t)^2}{2}\mathbf{g}
- (L \Delta t) \mathbf{v}
+ \boldsymbol{\Gamma}_{j} \mathbf{b}_a 
+ \lambda_i^n \boldsymbol{\mu}_i^n
- \lambda_j^n \boldsymbol{\mu}_j^n
= \mathbf{s}_{j} ~,
\end{split}
\label{eq:martinelli_constraint}
\vspace{-3pt}
\end{equation}
\noindent where the vector $\mathbf{s}_j$ and matrix $\boldsymbol{\Gamma}_{j}$ are computed by integrating the IMU measurements over the interval $[t_i, t_j]$ as
\vspace{-3pt}
\begin{equation}
        \mathbf{s}_j = \int_{t_i}^{t_j} (t_j - \tau) \, \R_{1\tau} \, \tilde\acc_\tau \, d\tau, \
            \boldsymbol{\Gamma}_{j} = \int_{t_i}^{t_j} (t_j - \tau) \, \R_{i\tau}  \, d\tau~.
\label{eq:matrix-gamma}
\vspace{-3pt}
\end{equation}

The rotation $\R_{i\tau} \in {SO}(3)$ comes from the IMU preintegration from time $t_i$ up to time $\tau$ using Eq.~\ref{eq:preintegrationRotation}. Note that $\bias^g$ was already initialized from the first two frames. The vector $\tilde\acc_\tau$ contains the accelerometer readings. 
By stacking the constraints from the $N$ tracked features from frame $i$ to frame $j$, we obtain the overdetermined linear system $\boldsymbol{\Xi} \, \mathbf{x} = \mathbf{s}$, which yields the following least-squares problem
\begin{mdframed}[linewidth=0.5pt, roundcorner=1pt,frametitleaboveskip=2pt,frametitlebelowskip=2pt, innertopmargin=0pt, innerbottommargin=3pt, frametitle={\small$\mathbf{g}, \, \mathbf{v}, \, {\bias^a}$ initialization},frametitlebackgroundcolor=gray!15]
\begin{equation}
    \arg \min_{\mathbf{x}} \left( \boldsymbol{\Xi} \, \mathbf{x} - \mathbf{s}\right)^2 \rightarrow \hat{\mathbf{x}} = \mathbf{v}_{\text{min}}\left( \boldsymbol{\Xi}^\top \boldsymbol{\Xi}\right)
    \label{eq:global-linear-system}
\end{equation}
\end{mdframed}
which we solve via singular value decomposition (SVD), being the optimal solution $\hat{\mathbf{x}}$ the eigenvector associated to the smallest eigenvalue $\mathbf{v}_{\text{min}}\left( \boldsymbol{\Xi}^\top \boldsymbol{\Xi}\right)$. Each row of the matrix $\boldsymbol{\Xi}$ corresponds to one instance of Eq.~(\ref{eq:martinelli_constraint}), leading to $3(M-1)N$ equations and $9+NM$ unknowns, with $M$ the number of images considered. $\boldsymbol{\Xi}$ denotes the coefficients matrix
\begin{equation*}
\rowcolors{0}{}{white}
\resizebox{\linewidth}{!}{$
\boldsymbol{\Xi} =
\left[
\begin{array}{cccccccccccc}
T_2 & -t_2 \mathbf{I}_3  & \Gamma_2 &  \mu_1^1 & \mathbf{0}_3 & \mathbf{0}_3 & -\mu_2^1 & \mathbf{0}_3 & \mathbf{0}_3 & \mathbf{0}_3 & \mathbf{0}_3 & \mathbf{0}_3\\
T_2 & -t_2 \mathbf{I}_3  & \Gamma_2 &   \mathbf{0}_3  & \mu_1^2 & \mathbf{0}_3 & \mathbf{0}_3  & -\mu_2^2 & \mathbf{0}_3 & \mathbf{0}_3 & \mathbf{0}_3 & \mathbf{0}_3\\
\vdots & \vdots & \vdots & \vdots & \vdots & \vdots & \vdots & \vdots & \vdots & \vdots & \vdots & \vdots\\
T_2 & -t_2 \mathbf{I}_3  & \Gamma_2 &   \mathbf{0}_3  & \mathbf{0}_3 & \mu_1^N & \mathbf{0}_3  & \mathbf{0}_3 & -\mu_2^N & \mathbf{0}_3 & \mathbf{0}_3 & \mathbf{0}_3\\
\vdots & \vdots & \vdots & \vdots & \vdots & \vdots & \vdots & \vdots & \vdots & \vdots & \vdots & \vdots\\
T_j & -t_j \mathbf{I}_3  & \Gamma_j & \mu_1^1 & \mathbf{0}_3 & \mathbf{0}_3 & \mathbf{0}_3 & \mathbf{0}_3 & \mathbf{0}_3 & -\mu_j^1 & \mathbf{0}_3 & \mathbf{0}_3\\
T_j & -t_j \mathbf{I}_3  & \Gamma_j & \mathbf{0}_3  &  \mu_1^2& \mathbf{0}_3 & \mathbf{0}_3 & \mathbf{0}_3 & \mathbf{0}_3 & \mathbf{0}_3  & -\mu_j^2 & \mathbf{0}_3\\
T_j & -t_j \mathbf{I}_3  & \Gamma_j & \mathbf{0}_3 & \mathbf{0}_3 & \mu_1^N & \mathbf{0}_3 & \mathbf{0}_3 & \mathbf{0}_3 & \mathbf{0}_3  & \mathbf{0}_3 & -\mu_j^N 
\end{array}
\right]~,
$}
\label{eq:xi-matrix}
\end{equation*}


\noindent with $ T_j = -\frac{\Delta t^2}{2} \mathbf{I}_3 $. 
The state to be initialized is defined as
\vspace{-3pt}
\begin{equation*}
\mathbf{x} = \left[ \mathbf{g}^\top, \, \mathbf{v}^\top, \, {\bias^a}^\top, \, \lambda_1^1, \, \ldots, \, \lambda_{1}^N, \, \ldots \, \lambda_{j}^1, \,  \ldots \, \lambda_{j}^N \right]^\top ,
\end{equation*}

\noindent and $\mathbf{s}$ contains the measurements obtained by Eq. (\ref{eq:matrix-gamma}) 
\begin{equation*}
\mathbf{s} = [\mathbf{s}_2^\top,\, \ldots, \, \mathbf{s}_2^\top, \, \mathbf{s}_3^\top, \, \ldots, \, \mathbf{s}_3^\top, \mathbf{s}_{j}^\top \, \ldots, \, \mathbf{s}_{j}^\top ]^\top.
\end{equation*}
This yields an initialization method that initializes $\R_{ij}$ and $\bias^g$ from just the first two frames, and the rest of the states when they become observable, as illustrated in pseudocode by the Algorithm~\ref{alg:viinit}. A detailed analysis of the observability criteria follows in the next subsection.
{Although feature depths appear in the linear formulation, they are treated as nuisance variables and not estimated recursively. Unlike filter-based methods such as OpenVINS~\cite{geneva2020openvins}, depths
are implicitly eliminated and never used beyond initialization. Our observability tests assess numerical conditioning of the reduced system,
rather than enforcing depth convergence.}

\begin{algorithm}[t]
\caption{\textbf{Visual-Inertial State Initialization}}
\label{alg:viinit}
\small
\begin{algorithmic}[1]

\Require Visual-inertial data;
         observability thresholds $\widebar{\Delta {z}}_{\text{th}}, \Delta \rho_{\text{th}}$

\Ensure Visual-inertial states $\mathbf{R}_{ij}, \mathbf{b}^g, \mathbf{t}_j, \mathbf{v}_j, \mathbf{b}_a, \mathbf{g}$
\State $\mathcal{Z}_i \gets \texttt{get\_first\_image\_and\_extract\_features}()$
\State $\mathcal{Z}_j \gets \texttt{get\_next\_image\_and\_track\_features}(\mathcal{Z}_i)$
\State $\{ \tilde{\boldsymbol\omega}_k \}, \{ \tilde{\mathbf{a}}_k \} \gets \texttt{get\_IMU\_readings}()$
\State $\R_{ij} \gets \texttt{get\_rotation}(\mathcal{Z}_i,\mathcal{Z}_j)$
\State $\mathbf{b}^g \gets \texttt{get\_bias}(\R_{ij})$ \Comment{Eqs.~\ref{eq:bg-commutative-model},\ref{eq:const-vel-model1} or \ref{eq:const-vel-model2}}
\State $\text{full\_obs} \gets \textbf{false}$ 

\While{$\text{full\_obs}$ = \textbf{false}}
    \State $\text{t\_obs} \gets \texttt{translation\_obs\_test}(\mathcal{Z}_i,\mathcal{Z}_j,\R_{ij}, \widebar{\Delta {z}}_{\text{th}})$
    \Statex \Comment{Eq.~\ref{eq:condition-parallax}}

    \If{$\text{t\_obs}$ = \textbf{true}}
        \State $\mathbf{s}_j \gets \texttt{integrate\_IMU}(\{ \tilde{\boldsymbol\omega}_k \}, \{ \tilde{\mathbf{a}}_k \}, \mathbf{b}^g)$
        \State $\text{full\_obs} \gets \texttt{full\_obs\_test}(\mathcal{Z}_i,\mathcal{Z}_j,\mathbf{s}_j, \Delta \rho_{\text{th}})$

\Statex \Comment{Eq.~\ref{eq:fullobstest}}
        \If{$\text{full\_obs}$ = \textbf{true}}
        
            \State $\mathbf{t}_j, \mathbf{v}_j, \mathbf{b}_a, \mathbf{g} \gets 
                   \texttt{get\_rest\_of\_vi\_states}(\mathbf{s}_j)$
                   \Statex \Comment{Eq.~\ref{eq:global-linear-system}}
                   
                   \State $\text{full\_obs} \gets \textbf{true}$
        \EndIf
    \EndIf

    \State $\mathcal{Z}_j \gets \texttt{get\_next\_image\_and\_track\_features}(\mathcal{Z}_i)$
    \State $\{ \tilde{\boldsymbol\omega}_k \}, \{ \tilde{\mathbf{a}}_k \} \gets \texttt{get\_IMU\_readings}()$
    \State $\mathbf{R}_{ij} \gets \texttt{get\_rotation}(\mathcal{Z}_i,\mathcal{Z}_j)$

\EndWhile

\end{algorithmic}
\end{algorithm}

\subsection{Two–stages Observability Test}

While camera rotation and gyroscope bias are observable from any two frames, and in practice we will initialize them from the first two in the sequence, the remaining visual–inertial states are not. Specifically, using the linear system defined in Eq.~(\ref{eq:global-linear-system}), the full state will only be observable and hence we will be able to initialize it when $\boldsymbol{\Xi}$ is full rank. Assessing this criterion at each frame, however, may be expensive, as the matrix $\boldsymbol{\Xi}$ grows quickly with the number of frames. Motivated by this, we will evaluate observability in two stages, that are described next.

\textbf{Translation Observability Test (Stage~1):}
This stage assesses the average parallax in the image plane between the \emph{first} frame at $t_i$ and a \emph{single} subsequent frame at $t_j$, advancing progressively over the image sequence until the criterion is met. 
Specifically, we first backproject each feature \(\mathbf{z}^{n}_i\) extracted at $t_i$ and project it at $t_j$ assuming rotation-only motion between the two camera frames
\begin{equation}
\hat{\mathbf{z}}^{n}_j\;=\;h\left(\mathtt{K}\R_{ij}^\top\mathtt{K}^{-1} \,\begin{bmatrix} {\mathbf{z}_i^n}^\top & 1 \end{bmatrix}^\top\right),
\end{equation}
where $h\left( \cdot \right)$ transforms the image points from homogeneous to Euclidean coordinates. The parallax between $t_i$ and $t_j$ will be reflected in the difference between the tracks predicted with just rotational motion $\hat{\mathbf{z}}^{n}_j$ and the real tracks ${\mathbf{z}}^{n}_j$. We aggregate the parallax of all features using their average:
\begin{equation}
\widebar{\Delta {z}}_j \;=\; \frac{1}{N}\sum\nolimits_{n=1}^{N}\big\|\,\mathbf{z}^{n}_j-\hat{\mathbf{z}}^{n}_j\big\|_2 \,,
\end{equation}

The condition to be met is expressed as
\begin{equation}
\widebar{\Delta {z}}_j \; > \; \widebar{\Delta {z}}_{\text{th}}, \quad \mathrm{(Translation~ Observability~Test)}
\label{eq:condition-parallax}
\end{equation}
If the condition is not met, we move to the next frame and repeat the test until the threshold is satisfied.

\textbf{Full Observability Test (Stage~2):}
This stage is triggered once the motion provides sufficient translational information, i.e., when the condition in Eq.~(\ref{eq:condition-parallax}) holds. Following~\cite{gnss2025}, 
we assess numerical observability via the Hessian $\mathbf{H}$, as follows
\begin{equation}
\mathbf{H}=\boldsymbol{\Xi}^\top \mathbf{\Sigma}^{-1}\boldsymbol{\Xi} \, ,
\end{equation}
\noindent with $\mathbf{\Sigma}$ the residual weights obtained following~\cite{camera2024}. Using the full $\mathbf{H}$ is misleading, as nuisance depths $\lambda_j^n$, are often ill-conditioned under low parallax, and prone to depressing some singular values. Hence, we marginalize depths via the Schur complement to obtain $\mathbf{H}^*$ over the nine states to be estimated (gravity, velocity, accelerometer bias) and evaluate the observability on this reduced matrix.

\begin{figure}[t] 
  \centering
\includegraphics[width=0.6\columnwidth]{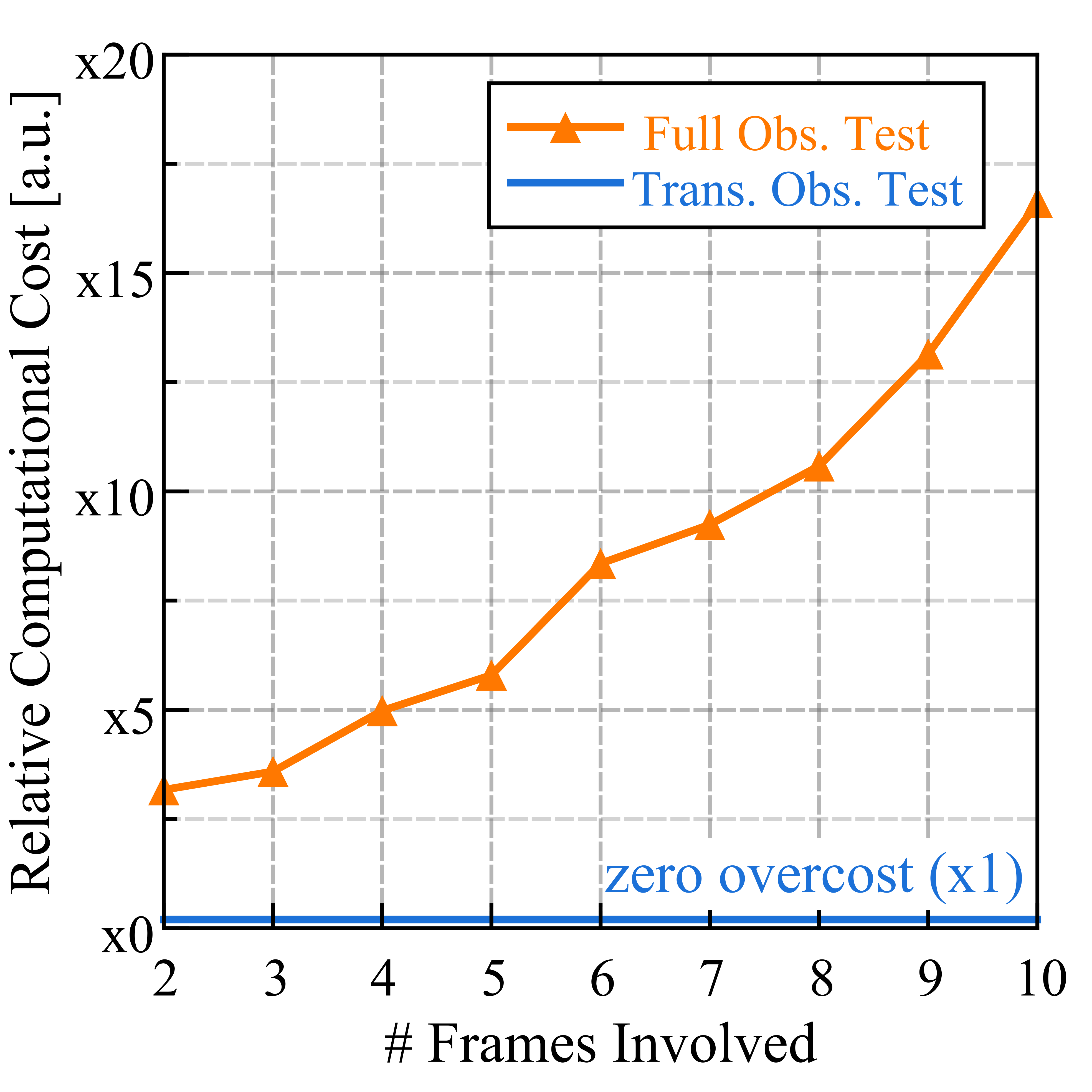}
\caption{\textbf{Relative computational costs of our observability tests}. Note the significantly higher cost of the full observability test, which motivates our proposal of a first stage assessing translational observability.}
  \label{fig:times-cond-comparison}
\end{figure}

We then compute the SVD of $\mathbf{H}^*\in\mathbb{S}^{9}_+$ and define the singular-value ratio $\rho_j=\sigma_{\max}/\sigma_{\min}$, where $\sigma_{\min}$ and $\sigma_{\max}$ denote the smallest nonzero and largest singular values, respectively. To track stabilization as the data window grows, we monitor the relative change
\begin{equation}    
\Delta \rho_j \;=\; \left| (\rho_j - \rho_{j-1})/\rho_{j-1} \right|,
\label{eq:fullobstest}
\end{equation}
\noindent where large values indicate that observability is still changing, whereas consistently small values indicate that the Hessian has stabilized and the state is observable. Once
\begin{equation}
    \Delta \rho_j \;<\; \Delta \rho_{\mathrm{th}}, \quad \mathrm{(Full~ Observability~Test) }
\end{equation}
with $\Delta \rho_{\mathrm{th}}$ a fixed threshold, we deem the system observable and proceed to solve Eq.~(\ref{eq:global-linear-system}). Fig.~\ref{fig:times-cond-comparison} reports the relative overcost of \textit{Full Observability Test} with respect to \textit{Translation Observability Test}.
The overcost of \textit{Full Observability Test} rises steadily with the window size: starting at roughly \(\!\times3\)–\(\!4\) the cost of \textit{Translation Observability Test} for short windows, but exceeding \(\!\times15\) around ten frames.
These results motivate the use of the parallax gate as the primary trigger and reserving the Hessian test as a final verifier.

\section{Experiments}
\subsection{Experimental Setup}
All experiments were conducted on a desktop PC equipped with an Intel Core i7-11700K (3.6\,GHz, 64\,GB RAM). 
Our method is implemented in C++, and nonlinear least-squares optimization is performed using Ceres~\cite{Ceres_Solver_2022}. 
Unless otherwise specified, all methods are executed on a single thread to ensure fair timing comparisons, and we report wall-clock times.
Performance is evaluated in terms of the root mean squared error (RMSE), expressed in the corresponding physical units.
{In line with previous research, we use the publicly available EuRoC MAV benchmark~\cite{EuRoc-dataset}, which provides a diverse set of motion profiles commonly used to benchmark
visual--inertial initialization methods, and is therefore adopted as our primary
real-world evaluation dataset.}
EuRoC provides synchronized 752\(\times\)480 stereo at 20\,Hz, inertial data at 200\,Hz and accurate ground truth. The dataset spans a range of motion profiles (easy/medium/difficult) and exhibits variations in texture, illumination, and dynamics, making it a suitable testbed for evaluating initialization accuracy, robustness, and runtime under realistic conditions. Following previous works~\cite{campos2020inertial, he2023rotation}, we perform exhaustive initialization every $0.5\,\mathrm{s}$ (i.e., every 10 frames) across all sequences, yielding a total of $2248$ independent initialization tests.

\subsection{Closed-form gyroscope bias}

\textbf{Synthetic data: }We conduct synthetic experiments in which the ground-truth relative rotations $\R_{ij}$ from EuRoC are progressively corrupted with zero-mean Gaussian noise, allowing us to isolate the effect of rotational uncertainty on the estimation accuracy of $\bias^g$. We report de-aggregated per-sequence metrics to assess generalization over different motion patterns.
{We compare the three closed-form approximation methods introduced in Section \ref{sec:rot-and-bias-estimation} with an iterative nonlinear least-squares optimization of $\bias^g$ based on Eq.~\eqref{eq:preintegrationRotation}.
All methods use the same measurements and relative rotations (estimated by C2P \cite{tirado2023correspondences}) within identical time windows selected by our observability-driven triggering strategy.}
As shown in Table~\ref{tab:bg-error-comparison-noise-free}, in an ideal noise-free setup, nonlinear optimization yields the lowest errors, while our closed-form solutions perform only slightly worse. However, as the noise level increases, the errors of our closed-form methods and the nonlinear baseline rapidly converge toward each other. Even for unrealistically small rotation errors of standard deviation 0.006$\,\mathrm{deg}$, the difference in $\bias^g$ error is already below $3\,\%$. For larger perturbations, the differences become negligible.
These results indicate that, in practical scenarios with noisy relative rotation estimations, our closed-form solutions achieve a $\bias^g$ estimation accuracy comparable to that of nonlinear optimization. Note also that the differences among our approximations are negligible.

\begin{table}[]
\centering
\begin{tabular}{l c  c c c }
\toprule
\multirow{2}{*}{Seq.} & \multirow{2}{*}{Nonlin. Opt.} & \multicolumn{3}{c}{Closed-form solutions (ours)} \\
& & Commutative & Avg. & Arithm. Avg. \\
\midrule
\multicolumn{5}{c}{\rule{0pt}{1.2em}\textbf{(a) Noise-free relative rotation}} \\
\rule{0pt}{1.2em}MH01 & \textbf{0.000002} & 0.000299 & 0.000042 & 0.000315 \\
MH02 & \textbf{0.000003} & 0.000303 & 0.000043 & 0.000329 \\
MH03 & \textbf{0.000004} & 0.000444 & 0.000055 & 0.000486 \\
MH04 & \textbf{0.000003} & 0.000327 & 0.000045 & 0.000338 \\
MH05 & \textbf{0.000003} & 0.000275 & 0.000042 & 0.000283 \\
V101 & \textbf{0.000003} & 0.000446 & 0.000052 & 0.000453 \\
V102 & \textbf{0.000004} & 0.000947 & 0.000104 & 0.000965 \\
V103 & \textbf{0.000007} & 0.001099 & 0.000123 & 0.001157 \\
V201 & \textbf{0.000004} & 0.000439 & 0.000059 & 0.000463 \\
V202 & \textbf{0.000004} & 0.000985 & 0.000109 & 0.001049 \\
V203 & \textbf{0.000008} & 0.001384 & 0.000166 & 0.001552 \\
\midrule
{Mean} & \textbf{0.000004} & 0.000631 & 0.000076 & 0.000672 \\
\midrule
\midrule
\multicolumn{5}{c}{\rule{0pt}{1.2em}\textbf{(b) Relative rotation noise of 0.006$\,\mathrm{deg}$ standard deviation}} \\
\rule{0pt}{1.2em}MH01 & \textbf{0.004148} & 0.004166 & 0.004149 & 0.004182 \\
MH02 & \textbf{0.004144} & \textbf{0.004144} & \textbf{0.004144} & 0.004185 \\
MH03 & \textbf{0.004148} & 0.004169 & \textbf{0.004148} & 0.004189 \\
MH04 & 0.004148 & 0.004147 & \textbf{0.004145} & \textbf{0.004145} \\
MH05 & 0.004148 & 0.004161 & \textbf{0.004147} & 0.004170 \\
V101 & 0.004148 & 0.004181 & \textbf{0.004147} & 0.004183 \\
V102 & \textbf{0.004147} & 0.004350 & 0.004158 & 0.004353 \\
V103 & \textbf{0.004147} & 0.004422 & 0.004163 & 0.004430 \\
V201 & \textbf{0.004148} & 0.004181 & \textbf{0.004148} & 0.004180 \\
V202 & 0.004149 & 0.004233 & \textbf{0.004144} & 0.004208 \\
V203 & \textbf{0.003711} & 0.004116 & 0.003729 & 0.004132 \\
\midrule
{Mean} & \textbf{0.004108} & 0.004206 & 0.004111 & 0.004214 \\
\midrule
\midrule
\multicolumn{5}{c}{\rule{0pt}{1.2em}\textbf{(c) Relative rotation noise of 0.06$\,\mathrm{deg}$ standard deviation}} \\
\rule{0pt}{1.2em}MH01 & \textbf{0.041481} & 0.041484 & 0.041482 & 0.041498 \\
MH02 & 0.041481 & \textbf{0.041462} & 0.041476 & 0.041502 \\
MH03 & \textbf{0.041481} & 0.041470 & \textbf{0.041481} & 0.041471 \\ 
MH04 & 0.041481 & 0.041462 & 0.041478 & \textbf{0.041458} \\
MH05 & 0.041481 & 0.041480 & \textbf{0.041479} & 0.041486 \\
V101 & 0.041481 & 0.041484 & \textbf{0.041480} & 0.041484 \\
V102 & \textbf{0.041481} & 0.041571 & 0.041490 & 0.041561 \\
V103 & \textbf{0.041481} & 0.041605 & 0.041495 & 0.041581 \\
V201 & \textbf{0.041481} & 0.041484 & \textbf{0.041481} & 0.041478 \\ 
V202 & 0.041483 & 0.041432 & 0.041477 & \textbf{0.041377} \\
V203 & 0.037116 & 0.037212 & 0.037129 & \textbf{0.037109} \\
\midrule
{Mean} & \textbf{0.041084} & 0.041104 & 0.041086 & 0.041091 \\
\midrule
\midrule
\multicolumn{5}{c}{\rule{0pt}{1.2em}\textbf{(d) Relative rotation noise from image matches}} \\
\rule{0pt}{1.2em}MH01 & 0.030048 & \textbf{0.030020} & 0.030046 & 0.030028 \\
MH02 & 0.033075 & \textbf{0.033052} & 0.033072 & 0.033064 \\
MH03 & \textbf{0.029123} & 0.029152 & 0.029126 & 0.029161 \\
MH04 & \textbf{0.045381} & 0.045410 & 0.045384 & 0.045392 \\
MH05 & 0.038635 & \textbf{0.038620} & 0.038633 & 0.038622 \\
V101 & 0.044986 & \textbf{0.044941} & 0.044980 & 0.044946 \\
V102 & 0.090967 & \textbf{0.090799} & 0.090948 & 0.090825 \\
V103 & 0.351947 & \textbf{0.351923} & 0.351939 & 0.351934 \\
V201 & 0.048773 & \textbf{0.048756} & 0.048769 & 0.048765 \\
V202 & \textbf{0.057977} & 0.058014 & 0.057978 & 0.058041 \\
V203 & 0.122305 & \textbf{0.122289} & 0.122303 & 0.122330 \\
\midrule
{Mean} & 0.081202 & \textbf{0.081180} & 0.081198 & 0.081192 \\
\bottomrule
\end{tabular}
\caption{{\textbf{Gyroscope bias errors [$\mathrm{rad/s}$] in EuRoC, for different noise setups for the relative rotation $\R_{ij}$}. \textbf{(a)} no noise, \textbf{(b)} standard deviation of 0.006$\,\mathrm{deg}$, \textbf{(c)} standard deviation of 0.06$\,\mathrm{deg}$, and \textbf{(d)} noise from image matches. We compare four methods: Nonlinear optimization (\textit{Nonlin. Opt.}), commutative approximation (\textit{Commutative.}), average approximation (\textit{Avg.}) and  arithmetic average approximation (\textit{Arithm. Avg.}). Best per row in \textbf{bold}. Except for the unrealistic noiseless case, all methods bring similar performance.}}
\label{tab:bg-error-comparison-noise-free}
\end{table}

\begin{figure*}[t]
    \centering
    \includegraphics[width=0.9\linewidth]{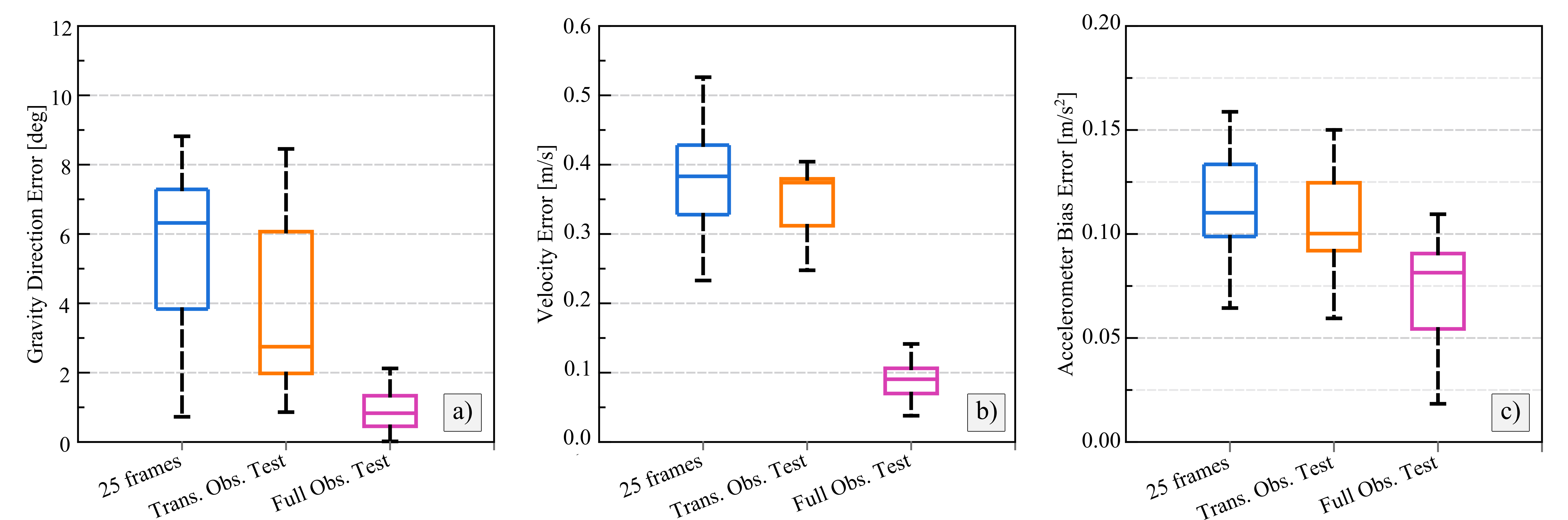}
\caption{\textbf{Initialization accuracy vs. activation policy.} We trigger closed-form initialization (i) after a fixed 25-frame window (\(\approx1.25\,\mathrm{s}\)); (ii) when \textit{Translation Observability Test} is satisfied; and (iii) when \textit{Full Observability Test} is satisfied. Panels report RMSE distributions in (a) gravity direction, (b) velocity, and (c) accelerometer bias.
The \textit{Translation Observability Test} already improves initialization accuracy, while the \textit{Full Observability Test} yields the best overall performance.
}
    \label{fig:triggering}
\end{figure*}

\textbf{Real data: }
In this experiment, relative rotations $\R_{ij}$ are computed from image correspondences using C2P~\cite{tirado2023correspondences}, with outliers removed via RANSAC. This setup emulates realistic conditions in which rotational inputs are derived from real visual data and are therefore affected by noise and matching inaccuracies.
Across all initialization experiments, the average outlier ratio was approximately $0.15$.

\begin{table}[h]
\centering
\begin{tabular}{lcc}
\toprule
{Method} & {Mean~[${\mu \mathrm{s}}$]} & {Std dev.~[${\mu \mathrm{s}}$]} \\
\midrule
Nonlinear optimization         & 3687 & 275 \\
Commutative approximation (ours)      & 94   & 6   \\
Avg. approximation (ours)  & 25  & 1   \\
\textbf{Arithm. Avg approximation (ours)} & \textbf{19}   & \textbf{1}   \\
\bottomrule
\end{tabular}
\caption{\textbf{Gyroscope bias initialization runtime.} Our approximations achieve up to two orders of magnitude speed-up compared to the nonlinear optimization baseline.}
\label{tab:bg_times}
\end{table}

The last rows in Table~\ref{tab:bg-error-comparison-noise-free} summarize the average gyroscope bias estimation errors across all EuRoC sequences, for our closed-form solutions and nonlinear optimization. 
Table~\ref{tab:bg_times} reports the mean and standard deviation of the execution time for gyroscope bias estimation under the four aforementioned strategies.
Relative to the nonlinear baseline, the \textit{Commutative} approximation reduces runtime by a factor $39\times$, the \textit{Avg.} approximation by $147\times$, and the \textit{Arithm. Avg.} approximation by $194\times$. This validates the computational advantage of our approximations. {Table~\ref{tab:bg_times} reports isolated gyroscope bias estimation runtimes, assuming that relative rotations $\R_{ij}$ are already available.
The computation of $\R_{ij}$ is included in the total initialization time reported in Table~\ref{tab:running-times}.}

As a summary, note how under realistic conditions all our methods achieve comparable accuracy, with noticeable differences only in the unrealistic noise-free case. The errors coming from feature matching are significantly higher that that of our approximations, which are negligible in practice.
The accuracy of our three proposed approximations is similar, while \textit{Arithm. Avg.} is the most efficient, and hence should be the preferred one for implementation.

\begin{table*}[t]
\centering
\setlength{\tabcolsep}{2.7pt}
\renewcommand{\arraystretch}{1.0}
\begin{tabular}{l|cccc|cccc|cccc|cccc}
\toprule
\multirow{2}{*}{{Seq.}}
& \multicolumn{4}{c|}{{Velocity Error~[$\mathrm{m/s}$]}}
& \multicolumn{4}{c|}{{Gravity Error~[$\mathrm{deg}$]}}
& \multicolumn{4}{c|}{{Accel. Bias Error~[$\mathrm{m/s^2}$]}}
& \multicolumn{4}{c}{{Gyro. Bias Error~[$\mathrm{rad/s}$]}} \\
\cmidrule(lr){2-5}\cmidrule(lr){6-9}\cmidrule(lr){10-13}\cmidrule(lr){14-17}
& \makecell{\small AS-MLE} & \makecell{\small VINS-\\[-2pt]Mono} & \makecell{\small DRT-\\[-2pt]tight} & \textbf{\small Ours}
& \makecell{\small AS-MLE} & \makecell{\small VINS-\\[-2pt]Mono} & \makecell{\small DRT-\\[-2pt]tight} & \textbf{\small Ours}
& \makecell{\small AS-MLE} & \makecell{\small VINS-\\[-2pt]Mono} & \makecell{\small DRT-\\[-2pt]tight} & \textbf{\small Ours}
& \makecell{\small AS-MLE} & \makecell{\small VINS-\\[-2pt]Mono} & \makecell{\small DRT-\\[-2pt]tight} & \textbf{\small Ours} \\
\midrule
MH01
& 0.11 & \textbf{0.08} & 0.11 & \underline{0.10}
& 1.64 & 1.17 & \underline{1.00} & \textbf{0.87}
& 0.093 & \underline{0.092} & -- & \textbf{0.088}
& 0.091 & \underline{0.070} & -- & \textbf{0.030} \\
MH02
& 0.09 & \textbf{0.08} & 0.11 & 0.09
& 1.36 & 1.11 & \underline{0.97} & \textbf{0.65}
& \underline{0.092} & 0.093 & -- & \textbf{0.091}
& 0.090 & \underline{0.080} & -- & \textbf{0.033} \\
MH03
& 0.30 & 0.19 & \underline{0.14} & \textbf{0.09}
& 3.06 & \underline{1.57} & \textbf{0.96} & \textbf{0.96}
& 0.080 & \underline{0.081} & -- & 0.080
& 0.063 & \underline{0.051} & -- & \textbf{0.029} \\
MH04
& 0.25 & \underline{0.14} & 0.19 & \textbf{0.11}
& 1.87 & 1.23 & \underline{0.99} & \textbf{0.84}
& \underline{0.087} & 0.088 & -- & \textbf{0.086}
& 0.068 & \underline{0.060} & -- & \textbf{0.045} \\
MH05
& 0.25 & \underline{0.17} & 0.22 & \textbf{0.13}
& 2.61 & 1.42 & \underline{0.99} & \textbf{0.83}
& \underline{0.084} & 0.081 & -- & 0.081
& 0.075 & \underline{0.062} & -- & \textbf{0.039} \\
\midrule
V101
& 0.07 & \textbf{0.06} & 0.07 & 0.07
& 3.37 & -- & \underline{0.99} & \textbf{0.89}
& \underline{0.316} & 0.319 & -- & \textbf{0.128}
& \underline{0.082} & -- & -- & \textbf{0.045} \\
V102
& 0.24 & 0.11 & \underline{0.06} & \textbf{0.05}
& 5.57 & 2.60 & \textbf{0.79} & \underline{1.09}
& 0.082 & \underline{0.081} & -- & \textbf{0.078}
& 0.091 & \textbf{0.077} & -- & {0.091} \\
V103
& -- & -- & \textbf{0.11} & \underline{0.12}
& -- & -- & \underline{1.48} & \textbf{0.87}
& \underline{0.115} & 0.116 & -- & \textbf{0.094}
& 0.352 & 0.349 & -- & {0.352} \\
V201
& 0.08 & \underline{0.07} & 0.08 & \textbf{0.06}
& 1.79 & 1.43 & \underline{1.15} & \textbf{1.06}
& 0.085 & \underline{0.083} & -- & \textbf{0.082}
& 0.067 & {0.049} & -- & {0.049} \\
V202
& 0.13 & 0.08 & \textbf{0.06} & \underline{0.07}
& 2.68 & 1.26 & \textbf{0.84} & \underline{0.94}
& \underline{0.055} & 0.056 & -- & \textbf{0.052}
& 0.068 & \underline{0.067} & -- & \textbf{0.058} \\
V203
& -- & -- & \underline{0.09} & \textbf{0.08}
& -- & -- & \underline{1.17} & \textbf{0.92}
& \textbf{0.053} & \textbf{0.053} & -- & {0.062}
& 0.151 & 0.141  & -- & \textbf{0.122} \\
\midrule
\textbf{Mean}
& 0.17 & 0.11 & 0.11 & \textbf{0.09}
& 2.66 & 1.47 & \underline{1.03} & \textbf{0.90}
& 0.104 & 0.104 & -- & \textbf{0.084}
& 0.109 & \underline{0.101} & -- & \textbf{0.081} \\
\bottomrule
\end{tabular}
\caption{\textbf{Per-sequence initialization errors on EuRoC (attempts every $0.5\,\mathrm{s}$)}
for AS-MLE, VINS-Mono, DRT-tight, and our method.
Best per row is shown in \textbf{bold}, second-best is \underline{underlined}.}
\label{tab:euroc-per-seq-rowwise}
\end{table*}

\subsection{Full state Initialization}
\textbf{Evaluation of Triggering strategies: }
To validate our observability-driven initialization, we compared three different activation strategies:  
(i) fixed-time policy after $25$ frames ($\approx 1.25~\mathrm{s}$), 
(ii) activation when the first observability criteria (\textit{Translation Observability Test}) is satisfied, and (iii) activation when the second observability criteria (\textit{Full Observability Test}) is satisfied. 
Fig.~\ref{fig:triggering} reports the results of this experiment and highlights the advantages of our two-stage strategy over a naive fixed-window policy. Initializing after $25$ frames yields noticeably larger and more dispersed errors, particularly in the gravity estimate, as no guarantees exist on the state observability in the chosen interval. By contrast, activating the closed-form full solution at the first \textit{Translation Observability Test} already achieves a clear reduction in error. And even greater improvements are obtained when monitoring the Hessian and postponing activation until the \textit{Full Observability Test} is satisfied, which consistently yields the lowest errors across all estimated state variables. These results confirm that our observability-driven criterion not only enables faster initialization than fixed-time approaches, but also provides more accurate and reliable estimates than a fixed-window strategy.
{The accelerometer bias is difficult to decouple from gravity under limited excitation.
Accordingly, Eq.~\eqref{eq:martinelli_constraint} is solved only after the Full Observability Test indicates a well-conditioned Hessian, as premature solving leads to biased estimates
(See Fig.~\ref{fig:triggering}).
Jointly estimating velocity, gravity, and $\mathbf{b}_a$ in a single linear system
provides additional numerical robustness without relying on 3D structure.}

\textbf{Comparison against baselines: }
We compare our initializer against three representative baselines: the
tightly-coupled version of DRT (denoted as DRT-tight)~\cite{he2023rotation},
the initialization module of VINS-Mono (denoted as VINS-Mono)~\cite{VINSmono},
and the analytical-solution method~\cite{zuniga2021analytical}, which extends
the ORB-SLAM3 initialization~\cite{campos2020inertial} (denoted as AS-MLE).
Table~\ref{tab:euroc-per-seq-rowwise} reports per-sequence initialization errors
on the EuRoC dataset. Averaged across all sequences, our method achieves the
lowest velocity RMSE ($0.09\,\mathrm{m/s}$) and gravity-direction error
($0.90\,\mathrm{deg}$), outperforming DRT-tight ($0.11\,\mathrm{m/s}$,
$1.03\,\mathrm{deg}$), VINS-Mono ($0.11\,\mathrm{m/s}$, $1.47\,\mathrm{deg}$),
and AS-MLE ($0.17\,\mathrm{m/s}$, $2.66\,\mathrm{deg}$). Per-sequence results
show that our initializer is best or second-best in the majority of cases for
both velocity and gravity, with only a few isolated sequences where DRT-tight
remains competitive (e.g., gravity estimation in V102). 
{In addition to velocity and gravity, Table~\ref{tab:euroc-per-seq-rowwise}
explicitly reports inertial bias errors during initialization.}
{Our method consistently achieves the lowest gyroscope bias error
across all sequences, with values in the order of $10^{-2}\,\mathrm{rad/s}$,
significantly improving over AS-MLE and VINS-Mono.}
{This highlights the benefit of explicitly estimating and compensating
the gyroscope bias \emph{prior} to solving the closed-form initialization system,
leading to more accurate relative rotations.}
{The improved rotation estimates directly translate into more consistent
gravity alignment and lower velocity errors, as observed in Table~\ref{tab:euroc-per-seq-rowwise}.
For the accelerometer bias, our approach also yields the lowest average RMSE
($0.084\,\mathrm{m/s^2}$), indicating that bias effects are reliably captured
during initialization.
We note that DRT-tight does not explicitly model inertial biases
during its initialization stage, and therefore accelerometer and gyroscope
bias errors are not reported for this method.
Unlike some initialization methods, we do not enforce a fixed gravity magnitude
during initialization.
Instead, translation-related states are estimated only after sufficient
observability is detected, ensuring a well-conditioned linear system.
As shown in Fig.~\ref{fig:triggering} and Table~\ref{tab:euroc-per-seq-rowwise}, this observability-aware strategy yields lower
gravity-direction and velocity errors than methods enforcing gravity magnitude
prematurely.}

\begin{table}[h]
\centering
\begin{tabular}{c| c c c c }
\toprule
Module & AS-MLE & VINS-Mono & DRT-tight & \textbf{Ours} \\
\midrule
SfM & 30.30 & 30.35 & -- & --\\
$\bias^g$ Est. & 0.15 & 0.44 & 1.95 & \textbf{0.02} \\
Vel. \& Grav. Est. & \textbf{0.08} & 0.14 & 2.81 & 0.52\\
Point Tri.  & \textbf{0.01} & \textbf{0.01} & 0.42 & 0.42 \\
\midrule
Total & 30.54 & 30.94 & 5.18 & \textbf{0.96}\\
\bottomrule
\end{tabular}
\caption{\textbf{Average initialization cost [$\mathrm{ms}$] on EuRoC}. Runtime breakdown and comparison against baselines.}
\label{tab:running-times}
\end{table}

\textbf{Runtime: }
We evaluate our overall initialization runtime against representative baselines on the EuRoC dataset.
In Table~\ref{tab:running-times} the full state initialization time is broken down by modules. 
Our method does not require 3D structure during initialization; therefore, we only include the cost of gyroscope bias estimation and the velocity–gravity solver. 
For a fair comparison with methods that rely on structure, we additionally report an optional ``Point Tri.'' stage executed after initialization. 
Our total initialization (0.96 $\mathrm{ms}$) is $31.8\times$ lower than AS-MLE, $32.2\times$ lower than VINS-Mono, and $5.4\times$ lower than DRT-tight. At the module level, gyroscope-bias estimation is reduced by a factor $7.5\times$ compared to AS-MLE, a factor $22\times$ compared to VINS-Mono, and a factor compared to DRT-tight. Our closed-form velocity–gravity solver is $5.4\times$ faster than DRT-tight, while being $3.7$–$6.5\times$ slower than AS-MLE/VINS-Mono (which leverage prebuilt structure). Notably, our pipeline avoids the SfM stage ($\approx30$ $\mathrm{ms}$ in AS-MLE/VINS-Mono), which dominates their cost; if reporting \emph{core-only} initialization (bias + vel. grav.), our time is $0.54$ $\mathrm{ms}$, i.e., $56.6\times$ lower than AS-MLE's.

\textbf{Initialization delay: }
Table~\ref{tab:init-windows-fixed} compares the initialization windows used by the baselines and our method. Prior approaches require temporal windows of roughly two seconds on average, either fixed or adaptively selected. In contrast, our two-stage criterion achieves reliable initialization with an adaptive window averaging only half a second.

\begin{table}[]
\centering
\begin{tabular}{lcc}
\toprule
{Method} & {{Initialization time [$\mathrm{s}$]}} & {Adaptive?}  \\
\midrule
Campos et al.~\cite{campos2019fast} & 2.16 $\pm$ 0.01 & \color{darkgreen}{\cmark} \\
He et al.~\cite{he2023rotation}     & 2.50   & \color{red}{\xmark} \\
\textbf{Ours}                    & \textbf{0.48} $\pm$ \textbf{0.27} & \color{darkgreen}{\cmark} \\
\bottomrule
\end{tabular}
\caption{\textbf{Initialization window length and adaptivity.} Whereas baselines rely on fixed/long windows, our two-stage criterion reduces initialization delays $4\times$.}
\label{tab:init-windows-fixed}
\end{table}

\begin{figure}[t]
\centering
\includegraphics[width=0.92\linewidth]{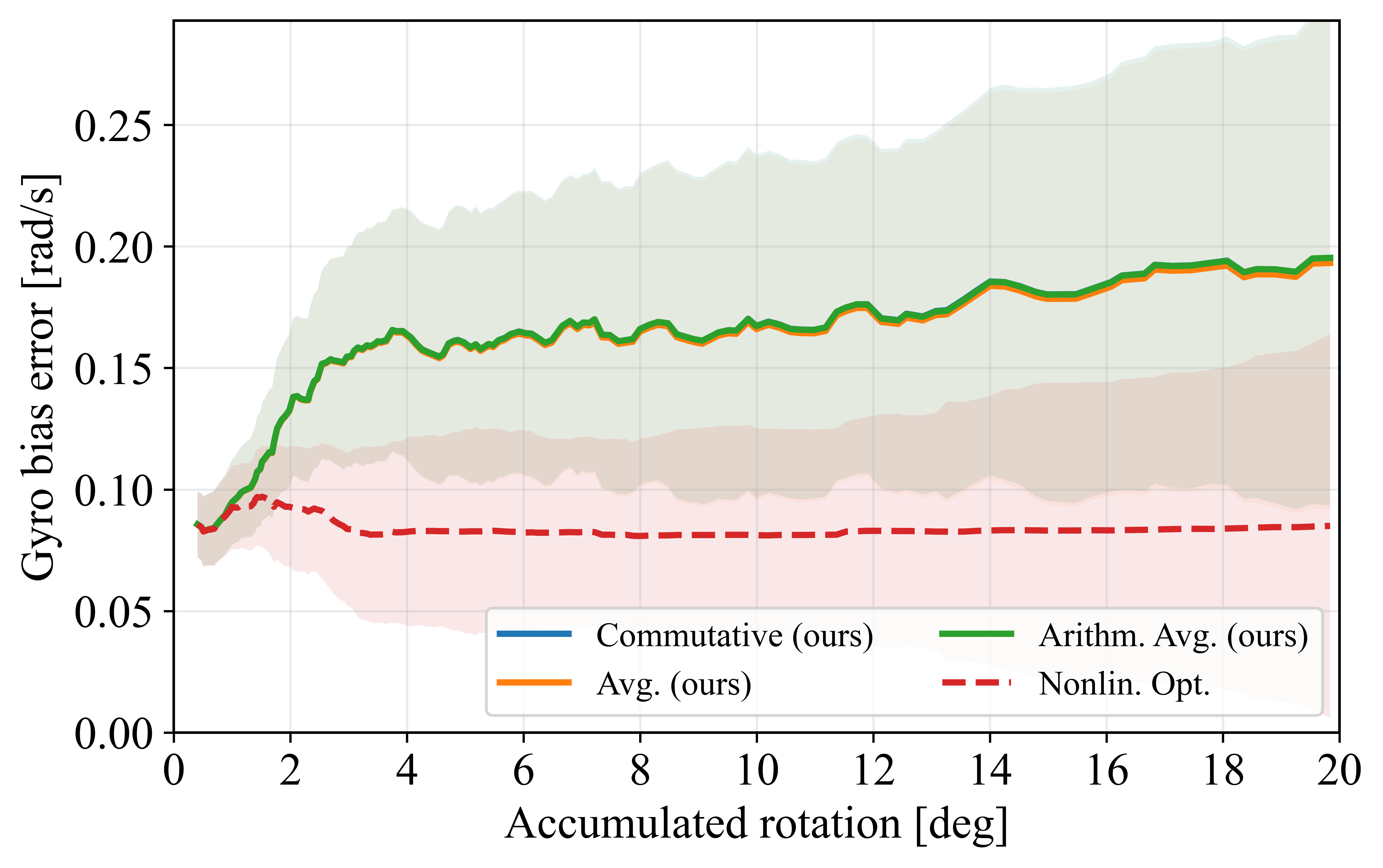}
\caption{\textbf{Gyroscope-bias error vs.\ accumulated rotation on EuRoC}. 
Our three approximation variants (solid blue/green/orange curves) produce nearly identical results and appear superimposed. 
In contrast, the nonlinear optimization (dashed red curve) remains comparatively stable up to 20$\,\mathrm{deg}$, while approximation-based estimates exhibit increasing error and variability as rotation grows.}
\label{fig:bgerr_vs_dtheta}
\end{figure}

\textbf{Approximation Robustness Analysis: }
To quantify the validity range of the three approximations, we plot in Fig.~\ref{fig:bgerr_vs_dtheta} the gyroscope bias estimation error as a function of the accumulated rotation magnitude.
For each valid sample, we compare the resulting bias error for: (i) the commutative approximation, (ii) the average approximation, (iii) the arithmetic average approximation and (ii) the nonlinear optimization.

Across 2698 valid samples, no sharp degradation is observed up to $1.5\!-\!2\,\mathrm{deg}$, where our approximations remains very accurate. Beyond this threshold, the error increases and the gap to nonlinear optimization becomes persistent. In contrast, nonlinear optimization remains stable, with a nearly flat trend and lower error. It should be noted, however, that only 15$\,\%$ of the consecutive pairs evaluated in the EuRoC dataset have relative rotations greater than 2$\,\mathrm{deg}$.
This observation supports the practical validity of our approximations and is consistent with the quantitative results reported in Table~\ref{tab:bg-error-comparison-noise-free} and Table~\ref{tab:euroc-per-seq-rowwise}.
\color{black}
\section{Conclusions}
\noindent In this work, we introduced a closed-form initialization that recovers
the full visual–inertial state without iterative solvers. By leveraging small-rotation and constant-velocity approximations, our methods preserves the essential couplings between motion and inertial measurements while
remaining analytically compact. A central ingredient is our fast approximation for gyroscope-bias estimation,
which replaces costly optimization with a lightweight solver and enables
immediate and robust start-up. Across all EuRoC sequences, our method reduces the initialization errors of relevant baselines by $10$--$20\%$, reduces initialization windows by $4\times$ and computation times by $5\times$. Importantly, our solutions are also algebraically simpler.
{Because the proposed approach is designed for short time windows and moderate motion, its underlying approximations may break down under highly dynamic conditions. In such occasional cases, slower nonlinear optimization-based methods may be more suitable.} Nevertheless, we believe our closed-form perspective offers a practical alternative for resource-constrained platforms and a strong foundation for future extensions
within modern visual-inertial pipelines.

\balance
\bibliographystyle{IEEEtran}
\bibliography{IEEEabrv,biblio}

\end{document}